# An Approach to Solve Linear Equations Using Time-Variant Adaptation Based Hybrid Evolutionary Algorithm

[1]A. R. M. Jalal Uddin Jamali, [2]M. M. A. Hashem, and [1]Md. Bazlar Rahman.

[1] Department of Mathematics
Khulna University of Engineering and Technology
Khulna-9203, Bangladesh
Ph: +88-041-769471 Ext-531, 523
**E-mail: jamali@math.kuet.ac.bd**

and
[2] Department of Computer Science and Engineering
Khulna University of Engineering and Technology
Khulna-9203, Bangladesh
Ph: +88-041-774318, Fax+88- 0141774403
E-mail: **hashem@cse.kuet.ac.bd**

***Abstract*** − For small number of equations, systems of linear (and sometimes nonlinear) equations can be solved by simple classical techniques. However, for large number of systems of linear (or nonlinear) equations, solutions using classical method become arduous. On the other hand evolutionary algorithms have mostly been used to solve various optimization and learning problems. Recently, hybridization of evolutionary algorithm with classical Gauss-Seidel based Successive Over Relaxation (SOR) method has successfully been used to solve large number of linear equations; where a uniform adaptation (UA) technique of relaxation factor is used. In this paper, a new hybrid algorithm is proposed in which a time-variant adaptation (TVA) technique of relaxation factor is used instead of uniform adaptation technique to solve large number of linear equations. The convergence theorems of the proposed algorithms are proved theoretically. And the performance of the proposed TVA-based algorithm is compared with the UA-based hybrid algorithm in the experimental domain. The proposed algorithm outperforms the hybrid one in terms of efficiency.

## 1. Introduction

Solving a set of simultaneous linear equations is a fundamental problem that occurs in diverse applications. A linear system can be expressed as a matrix equation in which each matrix or vector element belongs to a field, typically the real number system $\Re^n$. A set of linear equations in *n* unknown $x_1, x_2, \ldots, x_n$ and *n* number of equations is given by:

$$\sum_{j=1}^{n} a_{ij} x_j = b_i \, , \quad i = 1, 2, \cdots n \tag{1}$$

Equivalently, in matrix form:

$$\mathbf{Ax} = \mathbf{b} \, . \tag{2}$$

Where $\mathbf{A} = (a_{ij}) \in \Re^n \times \Re^n$ is the coefficient matrix, $\mathbf{x} = (x_j) \in \Re^n$ is the vector of unknown variables (i.e. solution vector), $\mathbf{b} = (b_i) \in \Re^n$ is the right hand constant vector.

Among the classical iterative methods, to solve system of linear equations, Gauss-Seidel based Successive Over Relaxation (SOR) method is one of the best methods. But the speed of convergence depends on over relaxation factor, $\omega$, with a necessary condition for the convergence being $0 < \omega < 2$ [1]. It is often very difficult to estimate the optimal relaxation factor and SOR technique is very sensitive to the relaxation factor, $\omega$ [2]. On the other hand the Evolutionary Algorithms (EA) are stochastic algorithms whose search methods model some natural phenomena: genetic inheritance and Darwinian strife for survival [3, 4, 5]. Almost all of the works on EA can be classified as evolutionary optimization (either numerical or combinatorial) or evolutionary learning. But Fogel and Atmar [6] used linear equation solving as test problems for comparing recombination, inversion operations and Gaussian mutation in an evolutionary algorithm. However, they emphasized their study not on equation solving, but rather on comparing the effectiveness of recombination relative to mutation. No comparison with classical equation-solving methods was given. Recently, a hybrid evolutionary algorithm [7] is developed by integrating classical Gauss-Seidel based SOR method with evolutionary computation techniques to solve equations in which $\omega$, is self-adapted by using Uniform Adaptation (UA) technique. For the cause of uniform adaptation, $\omega$'s are not finely adapted and there is a tendency of vibrating in any stages and there is no local fine-tuning in later stages. The idea of self-adaptation was also applied in many different fields [11].

Obvious biological evidence is that a rapid change is observed at early stages of life and a slow change is observed at latter stages of life in all kinds of animals/plants. These changes are more often occurred dynamically depending on the situation exposed to them. By mimicking this emergent natural evidence, a special dynamic Time-Variant Mutation (TVM) operator is proposed by Hashem [8] and Michalewicz et al. [9,10] in global optimization problems. In this paper, a new hybrid evolutionary algorithm is proposed in which a Time-Variant Adaptive (TVA) technique is introduced aiming at both improving the fine local tuning and reducing the disadvantage of uniform adaptation of relaxation factors as well as mutation. The proposed TVA-based hybrid algorithm does not require a user to guess or estimate the optimal relaxation factor. The proposed algorithm initializes uniform relaxation factors in a given domain and "evolves" it. The proposed algorithm integrates the Gauss-Seidel-based SOR method with evolutionary algorithm, which uses initialization, recombination, mutation, adaptation, and selection mechanisms. It makes better use of a population by employing different equation-solving strategies for different individuals in the population. Then these individuals can exchange information through recombination and the error is minimized by mutation and selection mechanisms. Experimental results show that the proposed TVA-based hybrid algorithm can solve linear equations within small time compared to the UA-based hybrid algorithm.

## 2. The Proposed Hybrid Algorithm

For the solution of the linear equations (1), Gauss-Seidel based SOR method [1] is given by:

$$x_i^{(k+1)} = x_i^{(k)} + \frac{\omega}{a_{ii}} \left( b_i - \sum_{j=1}^{i-1} a_{ij} x_j^{(k+1)} - \sum_{j=i}^{n} a_{ij} x^{(k)} \right), i = 1, 2, \cdots n \quad (3)$$

Now coefficient matrix **A** of the equation (2) can be decomposed as

$$\mathbf{A} = \mathbf{D} - \mathbf{L} - \mathbf{U} \quad (4)$$

Where $\mathbf{D} = (d_{ij})$ is the diagonal matrix, $\mathbf{L} = (l_{ij})$ is the lower triangular matrix and $\mathbf{U} = (u_{ij})$ is the upper triangular matrix. Then in matrix form equation (3) can be rewrite as:

$$\mathbf{x}^{(k+1)} = \mathbf{H}_\omega \mathbf{x}^{(k)} + \mathbf{V}_\omega \quad (5)$$

Where $\mathbf{H}_\omega = (\mathbf{I} - \omega \mathbf{D}^{-1}\mathbf{L})^{-1}\{(1-\omega)\mathbf{I} + \omega \mathbf{D}^{-1}\mathbf{U}\}$ and $\mathbf{V}_\omega = \omega(\mathbf{I} - \omega \mathbf{D}^{-1}\mathbf{L})^{-1}\mathbf{D}^{-1}\mathbf{b}$

Here, **I** is the identity matrix and $\omega \in (\omega_L, \omega_U)$ is the relaxation factor which influence the convergence rate of the SOR technique greatly; $\omega_L$ and $\omega_U$ are denoted as lower and upper boundary values of $\omega$.

Similar to many other evolutionary algorithms, the proposed hybrid algorithm always maintains a population of approximate solution to linear equations. Each solution is represented by an individual. The initial population is generated randomly form the field $\Re^n$. Different individuals use different relaxation factors. Recombination in the hybrid algorithm involves all individuals in a population. If the population size is N, then the recombination will have N parents and generate N offspring through linear combination. Mutation is achieved by performing one Gauss-Seidel based SOR iteration as given by (5). Initially $\omega$ is generated between $\omega_L(=0)$ and $\omega_U(=2)$ and then $\omega$ value is adapted stochastically during evolution. The fitness of an individual is evaluated based on the error of an approximate solution. For example, given an approximate solution (i.e., an individual) **z**, its error is defined by $\|e(\mathbf{z})\| = \|\mathbf{Az} - \mathbf{b}\|$. The relaxation factor is adapted after each generation, depending on how well an individual performs (in terms of error). The main steps of the TVA-based hybrid evolutionary algorithm is described as follows:

**Step 1: Initialization**

Generate, randomly from $\Re^n$, an initial population of approximate solutions to the linear equations (1) using different relaxation factor for each individual of the population. Denote the initial population as $\mathbf{X}^{(0)} = \{\mathbf{x}_1^{(0)}, \mathbf{x}_2^{(0)}, \ldots, \mathbf{x}_N^{(0)}\}$ where $N$ is the population size. Let $k \leftarrow 0$ where $k$ is the generation counter. And initialize corresponding relaxation factor $\omega$ as:

$$\omega_i = \begin{cases} \omega_L + \dfrac{d}{2} & \text{for } i = 1 \\ \omega_{i-1} + d & \text{for } 1 < i \leq N \end{cases} \quad (6)$$

Where $d = \dfrac{\omega_U - \omega_L}{N}$

**Step 2: Recombination**

Now generate $\mathbf{X}^{(k+c)} = \{\mathbf{x}_1^{(k+c)}, \mathbf{x}_2^{(k+c)}, \ldots, \mathbf{x}_N^{(k+c)}\}$ as an intermediate population through the following recombination:

$$\mathbf{X}^{(k+c)} = \mathbf{R}\left(\mathbf{X}^{(k)}\right)^t \quad (7)$$

Where $\mathbf{R} = (r_{ij})_{N \times N}$ is a stochastic matrix [12], and the superscript t denotes transpose.

**Step 3: Mutation**
Then generate the next intermediate population $\mathbf{X}^{(k+m)}$ from $\mathbf{X}^{(k+c)}$ as follows: For each individual $\mathbf{x}_i^{(k+c)}$ ($1 \leq i \leq N$) in population $\mathbf{X}^{(k+c)}$ produces an offspring according to Eqn. (5)

$$\mathbf{x}_i^{(k+m)} = \mathbf{H}_{\omega_i} \mathbf{x}_i^{(k+c)} + \mathbf{V}_{\omega_i}, \qquad i = 1, 2, \ldots, N. \tag{8}$$

Where $\omega_i$ is denoted as relaxation factor of the *i*th individual and $\mathbf{x}_i^{(k+m)}$ is denoted as *i*th (mutated) offspring, so that only one iteration is carried out for each mutation.

**Step 4: Adaptation**
Let $\mathbf{x}^{(k+m)}$ and $\mathbf{y}^{(k+m)}$ be two offspring individuals corresponding relaxation factors $\omega_x$ and $\omega_y$ and $\|e(\mathbf{x}^m)\|$ and $\|e(\mathbf{y}^m)\|$ are their corresponding errors (fitness value). Then the relaxation factors $\omega_x$ and $\omega_y$ are adapted as follows:

(a) If $\|e(\mathbf{x}^m)\| > \|e(\mathbf{y}^m)\|$, (i) then move $\omega_x$ toward $\omega_y$ by using

$$\omega_x^m = (0.5 + p_x)(\omega_x + \omega_y) \tag{9}$$

and (ii) move $\omega_y$ away from $\omega_x$ using

$$\omega_y^m = \begin{cases} \omega_y + p_y(\omega_U - \omega_y), & \text{when } \omega_y > \omega_x \\ \omega_y + p_y(\omega_L - \omega_y), & \text{when } \omega_y < \omega_x \end{cases} \tag{10}$$

Where $p_x = E_x \times N(0, 0.25) \times T_\omega$, an error decreasing stochastic parameter of **x**, and
$p_y = E_y \times |N(0, 0.25)| \times T_\omega$, an error decreasing search parameter of **y**.

Here, $N(0, 0.25)$ is the Gaussian distribution with mean 0 and standard deviation 0.25, $E_x$ = an exogenous parameter of **x**, $F_y$ = an exogenous parameter of **y**, $\omega^*$ is the optimal relaxation factor, $\omega_x^m$ & $\omega_y^m$ are adapted relaxation factors correspond to $\omega_x$ and $\omega_y$, and

$$T_\omega = (1 - \frac{t}{T})^\gamma, \text{ Time-variant adaptive parameter}, \tag{11}$$

Here, t = number of generation (iteration), T = maximum generation number, γ = an exogenous parameter of $T_\omega$.

(b) If $\|e(\mathbf{x}^m)\| < \|e(\mathbf{y}^m)\|$, then adapt $\omega_x$ and $\omega_y$ in the same way as above but reverse the order of $\omega_x^m$ and $\omega_y^m$.

(c) If $\|e(\mathbf{x}^m)\| = \|e(\mathbf{y}^m)\|$, no adaptation. So that $\omega_x^m = \omega_x$ and $\omega_y^m = \omega_y$.

**Step 5: Selection and Reproduction**
Select the best N/2 offspring individuals according to their fitness values (errors). Then reproduce of the above selected offspring (i.e. each parents individual generates two offspring). Then form the next generation of N individuals.

**Step 6: Termination**

If $min\{\|e(\mathbf{z})\| : \mathbf{z} \in \mathbf{X}\} < \eta$ (Threshold error), then stop the algorithm and get unique solution. If $min\{\|e(\mathbf{z})\| : \mathbf{z} \in \mathbf{X}\} \to \infty$, then stop the algorithm but fail to get any solution. Otherwise go to Step 2.

## 3. Convergence Theorems

The following theorem establishes the convergence of the hybrid algorithm.

**Theorem-1**: *If there exist an $\varepsilon$ $(0 < \varepsilon < 1)$ such that, for the norm of $\mathbf{H}_\omega$, $\|\mathbf{H}_\omega\| < \varepsilon < 1$, then $\lim\limits_{k \to \infty} \mathbf{x}^{(k)} = \mathbf{x}^*$, where $\mathbf{x}^*$ is the solution to the linear system of equations i.e., $\mathbf{Ax}^* = \mathbf{b}$.*

**Proof**: The individuals in the population at generation $k$ are $\mathbf{x}_i^{(k)}, i = 1, 2, \cdots N$. Let $\mathbf{e}_i^{(k)}$ be the error between the approximate solution $\mathbf{x}_i^{(k)}$ and exact solution $\mathbf{x}^*$ i.e. $\mathbf{e}_i^{(k)} = \mathbf{x}_i^{(k)} - \mathbf{x}^*$.

Then according to the recombination

$$\mathbf{x}_i^{(k+c)} = \sum_{j=1}^{N} r_{ij} \mathbf{x}_j^{(k)}, \quad i = 1, 2, \ldots, N$$

$$\therefore \|\mathbf{e}_i^{(k+c)}\| = \|\mathbf{x}_i^{(k+c)} - \mathbf{x}^*\| = \|\sum_{j=1}^{N} r_{ij} \mathbf{x}_j^{(k)} - \mathbf{x}^*\|$$

Since $\sum_{j=1}^{N} r_{ij} = 1$ and $r_{ij} \geq 0, \quad i = 1, 2, \ldots, N$

$$\therefore \|\mathbf{e}_i^{(k+c)}\| = \|\sum r_{ij} (\mathbf{x}_j^{(k)} - \mathbf{x}^*)\| \leq \sum \|r_{ij}(\mathbf{x}_j^{(k)} - \mathbf{x}^*)\| \leq \sum \|r_{ij}\| \|\mathbf{x}_j^{(k)} - \mathbf{x}^*\|$$

$$\therefore \|\mathbf{e}_i^{(k+c)}\| \leq \sum r_{ij} \|\mathbf{x}_j^{(k)} - \mathbf{x}^*\|$$

$$\therefore \|\mathbf{e}_i^{(k+c)}\| < \max\{\|\mathbf{e}_j^{(k)}\| : j = 1, \ldots, N\}$$

Again according to mutation for $i = 1, 2, \ldots N$

$$\mathbf{x}_i^{(k+m)} = \mathbf{H}_{\omega_i} \mathbf{x}_i^{(k+c)} + \mathbf{V}_{\omega_i}$$

and also since $\mathbf{x}^* = \mathbf{H}_{\omega_i} \mathbf{x}^* + \mathbf{V}_{\omega_i}$ then

$$\mathbf{x}_i^{(k+m)} - \mathbf{x}^* = \mathbf{H}_{\omega_i} (\mathbf{x}_i^{(k+c)} - \mathbf{x}^*)$$

$$\therefore \|\mathbf{x}_i^{(k+m)} - \mathbf{x}^*\| = \|\mathbf{H}_{\omega_i}(\mathbf{x}_i^{(k+c)} - \mathbf{x}^*)\| \leq \|\mathbf{H}_{\omega_i}\| \|(\mathbf{x}_i^{(k+c)} - \mathbf{x}^*)\| \leq \|\mathbf{H}_{\omega_i}\| \|\mathbf{e}_i^{(k+c)}\|$$

$$< \|\mathbf{H}_{\omega_i}\| \cdot \max\{\|\mathbf{e}_j^{(k)}\|; j = 1, \ldots N\}$$

$$\therefore \|\mathbf{e}_i^{(k+m)}\| < \varepsilon \cdot \max\{\|\mathbf{e}_j^{(k)}\|; j = 1, \ldots N\}$$

Now according to the selection mechanism, we have for $i = 1, \ldots N$

$$\|\mathbf{e}_i^{(k+1)}\| \leq \|\mathbf{e}_i^{(k+m)}\| < \varepsilon \cdot \max\{\|\mathbf{e}_j^{(k)}\|; j = 1, \ldots N\}$$

This implies that the sequence $\{\max\{\|\mathbf{e}_j^{(k+1)}\|; j = 1, \ldots N, \}; k = 0, 1, 2, \cdots\}$ is strictly monotonic decreasing and thus convergent.

The rate of convergence is accelerated by TVA-based hybrid method. The following theorem justifies the adaptation technique for relaxation factors used in proposed TVA-based hybrid evolutionary algorithm.

**Theorem –2:** *Let $\rho(\omega)$ be the spectral radius of matrix $\mathbf{H}_\omega$ and let $\omega^*$ be the optimal relaxation factor, and $\omega_x$ and $\omega_y$ be the relaxation factors of the selected pair individuals $x$ and $y$ respectively. Assume $\rho(\omega)$ is monotonic decreasing when $\omega < \omega^*$ and $\rho(\omega)$ is monotonic increasing when $\omega > \omega^*$. Also consider $\rho(\omega_x) > \rho(\omega_y)$. Then*

(i) $\rho(\omega_x^m) < \rho(\omega_x)$, when $\omega_x^m = (0.5 + p_x)(\omega_x + \omega_y)$ and

(ii) $\rho(\omega_y^m) < \rho(\omega_y)$, when $\omega_y^m = \omega_y + p_y(\omega_U - \omega_y)$    if $\omega_y > \omega_x$
$\qquad\qquad\qquad\qquad\qquad\qquad = \omega_y + p_y(\omega_L - \omega_y)$    if $\omega_y < \omega_x$

**Proof**: We first assume that $\rho(\omega)$ is monotonic decreasing when $\omega < \omega^*$ and $\rho(\omega)$ is monotonic increasing when $\omega > \omega^*$. Let $\omega_x$ and $\omega_y$ be the two relaxation factors of a randomly selected pair individuals **x** and **y** and also let $\rho(\omega_x) > \rho(\omega_y)$. Then there will be four cases:

**Case-1** Both $\omega_x, \omega_y < \omega^*$: Since $\rho(\omega)$ is monotonic decreasing when $\omega < \omega^*$ and as assume $\rho(\omega_x) > \rho(\omega_y)$; so $\omega_x < \omega_y$. Then we get $\omega_x < \omega_x^m < \omega_y$ and so $\rho(\omega_x^m) < \rho(\omega_x)$, where $\omega_x^m = \omega_x + (0.5 + p_x)(\omega_x + \omega_y)$. Again since $\omega_x < \omega_y$, so we get $\omega_y < \omega_y^m \leq \omega^*$ and therefore $\rho(\omega_y^m) < \rho(\omega_y)$ where $\omega_y^m = \omega_y + p_y(\omega_U - \omega_y)$.

**Case-2** Both $\omega_x, \omega_y > \omega^*$: Since $\rho(\omega)$ is monotonic increasing when $\omega > \omega^*$ and as assume $\rho(\omega_x) > \rho(\omega_y)$; so $\omega_x > \omega_y$. Then we get $\omega_x > \omega_x^m > \omega_y$ and so $\rho(\omega_x^m) < \rho(\omega_x)$, where $\omega_x^m = \omega_x + (0.5 + p_x)(\omega_x + \omega_y)$. Again since $\omega_x > \omega_y$, so we get $\omega^* \leq \omega_y^m < \omega_y$ and therefore $\rho(\omega_y^m) < \rho(\omega_y)$, where $\omega_y^m = \omega_y + p_y(\omega_L - \omega_y)$.

**Case-3** If $\omega_x < \omega^* < \omega_y$: Since $\rho(\omega)$ is monotonic decreasing when $\omega < \omega^*$ and as assume $\rho(\omega_x) > \rho(\omega_y)$; so we get $\omega_x < \omega_x^m < \omega^* < \omega_y$ and therefore $\rho(\omega_x^m) < \rho(\omega_x)$, where $\omega_x^m = \omega_x + (0.5 + p_x)(\omega_x + \omega_y)$. Again since $\omega_x < \omega^* < \omega_y$ so we get $\omega^* < \omega_y < \omega_y^m$ and therefore $\rho(\omega_y^m) > \rho(\omega_y)$, where $\omega_y^m = \omega_y + p_y(\omega_U - \omega_y)$.

[Note that in this case $\omega_y^m$ will go a bit away from $\omega^*$ with a very small probability]

**Case-4** If $\omega_y < \omega^* < \omega_x$ : Since $\rho(\omega)$ is monotonic increasing when $\omega > \omega^*$ and as assume $\rho(\omega_x) > \rho(\omega_y)$; so we get $\omega_y < \omega^* < \omega_x^m < \omega_x$ and therefore $\rho(\omega_x^m) < \rho(\omega_x)$, where $\omega_x^m = \omega_x + (0.5 + p_x)(\omega_x + \omega_y)$. Again since $\omega_y < \omega^* < \omega_x$ so we get $\omega_y^m < \omega_y < \omega^*$, and therefore $\rho(\omega_y^m) > \rho(\omega_y)$, where $\omega_y^m = \omega_y + p_y(\omega_L - \omega_y)$.

[Note that in this case $\omega_y^m$ will go a bit away from $\omega^*$ with a very small probability]

If $\|e(\mathbf{x}^m)\| < \|e(\mathbf{y}^m)\|$, adapt $\omega_x$ and $\omega_y$ in the same way as above but reverse the role of $\omega_x^m$ and $\omega_y^m$. Considering all the above situations, we may conclude that the comparatively worse relaxation factor $\omega_x$ is always improved and the comparatively better relaxation factor $\omega_y$ also is improved with a very high probability in each generation. Hence, the proposed time-variant based adaptation mechanism increase the rate of convergence.

## 4. Numerical Experiment

In order to evaluate the effectiveness of the proposed TVA-based hybrid algorithm, numerical experiment have been carried out on a number of problems to solve the systems of linear Eqn. (2) of the form: $\mathbf{Ax} = \mathbf{b}$.

First we solve the problem of the above equation by setting the parameters: $a_{ii} = 2n$, and $b_i = i$ for $i = 1, 2, \cdots, n$ and $a_{ij} = j$ for $i \neq j$ $i, j = 1, 2, \cdots, n$; the dimension of the coefficient matrix $\mathbf{A}$ is $100 \times 100$ (i.e. $n = 100$). The population size $N$ is set at 2; $\omega_U$ and $\omega_L$ are set at 2 and 0 respectively so that initial $\omega$'s become 0.5 and 1.5. The exogenous parameters $\gamma$, $E_x$ and $F_y$ are set at 40.0, 0.1 and 0.01 respectively and maximum generation, T, is set at 2500. Each individual is initialized from the domain $\Re^{100} \in (-30, 30)$ randomly and uniformly. The problem is to be solved with an error smaller than $10^{-13}$ (threshold error). The program is run 10 times using 10 different sample paths and then averaged them. Figure 1 shows the numerical results (in graphical form) achieved by the UA-based hybrid algorithm [7] and proposed TVA-based hybrid algorithm. It is observed in Fig. 1 that the rate of convergence of TVA-based algorithm is much better than that of UA-based algorithm and TVA-based algorithm exhibits a fine local tuning.

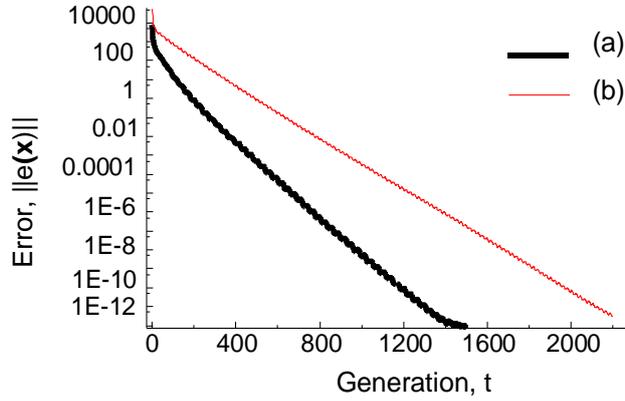

Figure 1: Curve (a) represents the Evolution History of TVA-based hybrid algorithm and curve (b) represents the Evolution History of UA-based hybrid algorithm

Table I shows six test problems, labeled from $P_1$ to $P_6$, with dimension, $n = 100$. For each test problem $P_i$: $i = 1, 2 \ldots 6$, the coefficient matrix $\mathbf{A}$ and constant vector $\mathbf{b}$ are all generated uniformly and randomly within given domains (shown in 2nd column with corresponding rows of Table I) and also initial population $\mathbf{X}$ are all generated uniformly and randomly within domain [-30,30]. Initial relaxation factors are set at 0.5 and 1.5 for all the cases. For different problems ($P_1$–$P_6$) corresponding threshold errors, $\eta$, and corresponding maximum generations (iterations), T, are shown in the Table I.

**Table I**
Comparison between TVA-based and UA-based algorithms for several randomly generated test problems

| Label of Test Problems | Domain of the Parameters of the test Problems | Threshold Error, $\eta$ | Maximum Generation, T | For UA-based | | For TVA-Based | |
|---|---|---|---|---|---|---|---|
| | | | | * Elapsed Time (in Second) | Generation (elapsed) | * Elapsed Time (in Second) | Generation (elapsed) |
| $P_1$ | $a_{ii}= 2n$; $a_{ij}=j$; $b_i= i$ | $10^{-12}$ | 2000 | 107 | 1812 | 60 | 910 |
| $P_2$ | $a_{ii}\in(-70,70)$; $a_{ij}\in(-2,2)$ $b_i\in(-2,2)$ | $10^{-12}$ | 500 | 18 | 297 | 06 | 108 |
| $P_3$ | $a_{ii}\in(-70,70)$; $a_{ij}\in(0,4)$ $b_i\in(0,70)$ | $10^{-12}$ | 500 | 29 | 465 | 27 | 434 |
| $P_4$ | $a_{ii}\in(1,100)$; $a_{ij}\in(-2,2)$ $b_i = 2$ | $10^{-12}$ | 1500 | 75 | 1260 | 39 | 445 |
| $P_5$ | $a_{ii} = 200$; $a_{ij}\in(-30,30)$ $b_i\in(-400,400)$ | $10^{-11}$ | 700 | 35 | 596 | 21 | 359 |
| $P_6$ | $a_{ii}\in(-70,70)$; $a_{ij}\in(0,4)$ $b_i\in(0,70)$ | $10^{-09}$ | 6000 | 350 | 5719 | 191 | 3236 |

* Elapsed Times are shown, in column five and in column eight, just for relative comparison of two algorithms.
Note: 1. The elements of coefficient matrix **A**, **b**, and initial population **X** are identical for each comparison.
2. Algorithms are implemented in Borland C++ environment using Pentium IV PC (1.2GHz).

Table I shows the comparison of the number of generation (iteration) and relative elapsed time used by the UA-based hybrid algorithm and by proposed TVA-based hybrid algorithm to the given preciseness, $\eta$ (see column three of the Table I). One observation can be made immediately from the Table I, except for problem $P_3$ where the UA-based method performed near to same as TVA-based method, TVA-based hybrid algorithm performed much better than the UA-based hybrid algorithm for all other problems.

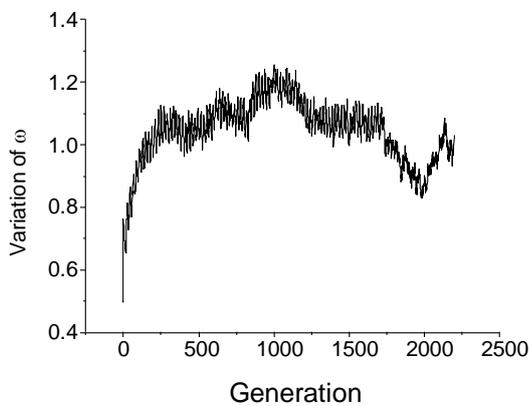
Figure 2: Self-adaptation of $\omega$ in the UA-based Algorithm

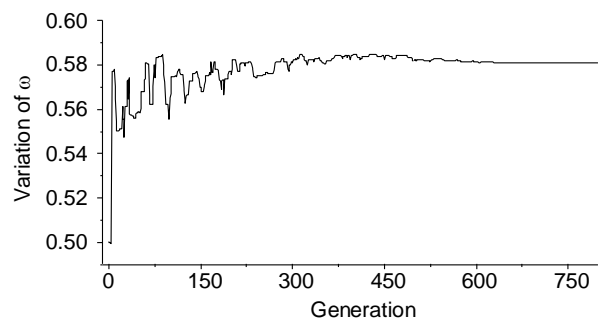
Figure 3: Self-adaptation of $\omega$ in the TVA-based algorithm

Fig. 2 shows the nature of self-adaptation of $\omega$ in the UA-based hybrid algorithm and Fig. 3 shows the nature of self-adaptation of $\omega$ in the TVA-based hybrid algorithm. It is observed in Fig. 2 and Fig. 3 that the self-adaptation process of relaxation factors in TVA-based hybrid algorithm is much better than that of in UA-based hybrid algorithm. Fig. 3 shows that how initial $\omega = 0.5$, is adapted to its near optimum value and reaches to a best position for which rate of convergence is accelerated. On the other hand Fig. 2 shows that initially $\omega = 0.5$, by self-adaptation process, does not gradually reaches to a best position. It is noted that both UA-based and TVA-based algorithms are implemented in Borland C++ environment.

## 5. Concluding Remarks

In this paper, a Time-variant adaptive (TVA)-based hybrid evolutionary algorithm has been proposed for solving linear systems of equations. The TVA-based hybrid algorithm integrates the classical Gauss-Seidel based SOR method with evolutionary computation techniques. The time-variant based adaptation is introduced for adaptation of relaxation factors, which makes the algorithm more natural and accelerates its rate of convergence. The recombination operator in the algorithm mixed two parents by a kind of averaging, which is similar to the intermediate recombination often used in evolution strategies [4, 5]. The mutation operator is equivalent to one iteration in the Gauss-Seidel based SOR method. The mutation is stochastic since the relaxation factor $\omega$ is adapted stochastically. The proposed TVA-based relaxation factor $\omega$ adaptation technique acts as a local fine tuner and helps to escape from the disadvantage of uniform adaptation. Numerical experiments with the test problems have shown that the proposed TVA-based hybrid algorithm performs better than the UA-based hybrid algorithm. Also TVA-based hybrid algorithm is more efficient and robust than the UA-based hybrid algorithm.